\documentclass[letterpaper, 10 pt, conference]{ieeeconf}  

\IEEEoverridecommandlockouts                              

\usepackage{graphics} 
\usepackage{epsfig} 
\usepackage{mathptmx} 
\usepackage{times} 
\usepackage{amsmath} 
\usepackage{amssymb}  
\usepackage{array}
\usepackage{hyperref}
\newcolumntype{H}{>{\centering\arraybackslash}m{2cm}}
\newcolumntype{L}{>{\centering\arraybackslash}m{6cm}}

\title{\LARGE \bf A Generalized Mixed-Integer Convex Program for Multilegged Footstep Planning on Uneven Terrain}

\author{Bernardo Aceituno-Cabezas, \textit{Student Member, IEEE}, Jos{\'e} Cappelletto,\\ Juan C. Grieco, and Gerardo Fern{\'a}ndez-L{\'o}pez, \textit{Member, IEEE}
\thanks{This research was partially funded by the Universidad Sim{\'o}n Bol{\'i}var Research and Development Deanship}
\thanks{B. Aceituno-Cabezas, J. Cappelletto, J. C. Grieco and G Fern{\'a}ndez-L{\'o}pez are with the Mechatronics Research Group, Sim{\'o}n Bol{\'i}var University, 1080 Sartenejas, Venezuela
	{\tt \small \{12-10764, cappelletto, jcgrieco, gfernandez\}@usb.ve}}%
}

\begin{document}

\maketitle
\thispagestyle{empty}
\pagestyle{empty}

\begin{abstract}
Robot footstep planning strategies can be divided in two main approaches: discrete searches and continuous optimizations. While discrete searches have been broadly applied, continuous optimizations approaches have been restricted for humanoid platforms. This article introduces a generalized continuous-optimization strategy for multilegged footstep planning which can be adapted to different platforms, regardless the number and geometry of legs. This approach leverages \textit{Mixed-Integer Convex Programming} to account for the non-convex constraints that represent footstep rotation and obstacle avoidance. The planning problem is formulated as an optimization problem, which considers robot geometry and reachability with linear constraints and can be efficiently solved using optimization software. To demonstrate the functionality and adaptability of the planner, a set of tests are performed on a BH3R hexapod and a LittleDog quadruped on scenarios which can't be easily handled with discrete searches, such tests are solved efficiently in fractions of a second. This work represents, to the knowledge of the authors, the first successful implementation of a continuous optimization-based multilegged footstep planner. 
\end{abstract}


\section{INTRODUCTION}	

Motion planning for walking robots has become an active area of research on the recent years. One particular and well-established method is based in the separation of the contact sequences and body motion plans as different problems. This consists in specifying a sequence of contacts in the form of footsteps for the robot to follow, and then solve for body motions separately. Such method requires powerful planning tools to obtain optimal and feasible footstep plans for the platform. As noted by Deits and Tedrake \cite{deits2014footstep}, footstep planning approaches can be broadly divided in two areas: \textit{discrete searches and continuous optimizations}. While discrete searches have been thoroughly studied for multilegged robots, the entirety of continuous optimization approaches have been restricted to bipedal platforms. The main issue when adapting such approach to multilegged platforms, is that representing the geometry of a multilegged robot and its workspace in a continuous optimization footstep planner results in the introduction of non-convex constraints. This results in a nonlinear program which is difficult to solve and can't guarantee an optimal result.

In the past, several graph-search algorithms such as A* or RRT have been successfully applied to multilegged footstep planning: Zucker et al. \cite{zucker2010optimization} use ARA* with learned cost maps to obtain plans for a LittleDog robot walking on challenging terrain, Satzinger et al. \cite{satzinger2015tractable} introduced a planner based on a modified version of A* to obtain footstep plans for a quadruped platform given a pre-computed foothold set to follow during the DARPA Robotics Challenge (DRC) finals, Winkler et al. \cite{winkler2015planning} build cost maps based on the complexity of terrain, and then apply ARA* to obtain footsteps for a Quadruped. However, these methods require \textit{pre-computed foothold sets} and terrain cost maps in order to provide a good heuristic for the search. These are difficult to compute without a prior knowledge of the desired path and are limited to a \textit{discrete} version of the environment.

On the other hand, continuous optimization approaches have been restricted for humanoid platforms. Herdt et al. \cite{herdt2010online} introduced a Quadratic Programming (QP) formulation that can produce optimal footstep placements and controls for a walking robot model when footstep orientations are fixed and obstacles in the environment are ignored. Fallon et al. \cite{fallon2015architecture} implemented a nonlinear programming algorithm that can plan footsteps while considering orientation, but couldn't guarantee optimality, nor generate plans around obstacles. Deits and Tedrake \cite{deits2014footstep} generalized the previous approach by formulating the problem as a \textit{Mixed-Integer Quadratically Constrained Quadratic Program} (MIQCQP) for the DRC Finals, constraining every step to lie within a safe convex region, and using a \textit{piecewise linear approximation} of the $sin$ and $cos$ functions to plan orientations. Such approach follows a similar idea as the work of Richards et al. \cite{richards2002coordination}, where safe UAV trajectories are obtained by using binary variables to restrict the plan within a set of \textit{safe convex regions} represented as the intersection of the half-planes of the obstacles faces.

This paper introduces a generalized continuous optimization footstep planner for multilegged platforms, which can be formulated as a \textit{Mixed-Integer Quadratic Program} (MIQP), solved efficiently using commercial convex optimization software, and can be easily adapted to multilegged robots with different geometries. The remaining of this paper is organized as follows: Section II describes the formulation of the program and its constraints, Section III presents the results obtained from a implementation of the planner in two different multilegged platforms (a LittleDog quadruped and a BH3R circular hexapod) and Section IV discusses and concludes on the contributions of this work.

\section{APPROACH}

The goal of the planner is to solve for a sequence of $N$ footsteps represented as:

\begin{equation}
	f \ = \ (
		f_{x} \
		f_{y} \
		f_{z} \
		\theta)
\end{equation}

where $f_{x}$, $f_{y}$, $f_{z}$ and $\theta$ are the $xyz$ coordinates of the foot and the approximated yaw angle of the body for that step. The planner is then formulated as a continuous optimization problem over the footstep sequence, subject to the constraints that: 

\begin{enumerate}
	\item Every footstep must be reachable from the previous position of the foot.
    \item All Footsteps must lie within a obstacle-free convex region.
\end{enumerate}

Where the geometry of the robot and the safe convex regions are given.

\subsection{Kinematic Reachability}

Since multilegged robots can have very different geometries, which determines the configurations that can be reached by each foot and the distribution of the leg workspace, it is necessary to restrict each step according such geometry. For this, every step is constrained around a nominal position. The \textit{Center of Contacts} (CoC or $p$) of a footstep is defined as the geometric center of each $n_{legs}$ configuration of footsteps finishing in such step, algebraically:

\begin{equation}
p_i = \frac{\sum_{k = i - n_{legs} + 1}^{i}{f_k}}{n_{legs}}
\end{equation}

Then, given the CoC for a certain footstep and a nominal distance $L_{leg}$ between both, the nominal leg position $r_{nom}$ is defined as:

\begin{equation}
r_{nom} = \begin{pmatrix} p_{x} + L_{leg} cos(\phi) \\ p_{y}  + L_{leg} sin(\phi) \end{pmatrix}
\end{equation}

\begin{figure}[ht!]
	\centering
	\includegraphics[width=2.5in]{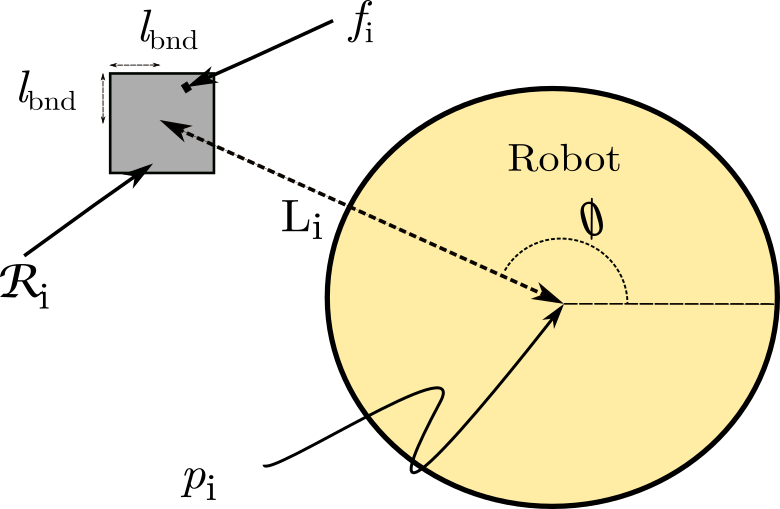}
	\caption{Geometric constraint in $f_i$ and $\Re$}
	\label{fig:fig3}
\end{figure}

Where $\phi$ is the angle between the approximate body orientation $\theta$ and the nominal leg position. Then, the step is required to land within a reference region $\Re$ centered on the reference position. In this case the region is chosen to be square shaped, making it rotation invariant in the trajectory (as shown in figure \ref{fig:fig3}):

\begin{equation}
f_\imath \in \Re \Leftrightarrow f_\imath \in \begin{pmatrix} r_{nomx} \pm l_{bnd} \\ r_{nomy} \pm l_{bnd} \end{pmatrix}
\end{equation}

Where $l_{bnd}$ is the side of the square that bounds the reference region.

The presence of the $sin$ and $cos$ functions in equation (3) makes the problem non-convex. In order to solve this problem, these functions are represented using a piecewise linear approximation, thus maintaining the convexity of the problem.

It is important to remark that the nominal length $L_{leg}$ should vary when the roll and pitch angles of the body differ significantly from 0. One solution is to represent this variable with the difference of height on the $n_{legs}$ footstep configuration as a bilinear constraint, which can be approximated with McCormick Envelopes or Multiparametric Disaggregation Techniques. In this work $L_{leg}$ is fixed as a constant, making the geometric constraints linear.

For a feasible plan, every footstep needs to be reachable from its previous configuration. Since the workspace of each leg often results in a non-convex set, it is necessary to approximate it in a way such that it can be represented as a convex constraint.  For this, every footstep is restricted to lie in a polygon inscribed within the biggest circle centered at $r_{nom}$ and contained in the workspace.

\begin{figure}[ht!]
\centering
\includegraphics[width=3.0in]{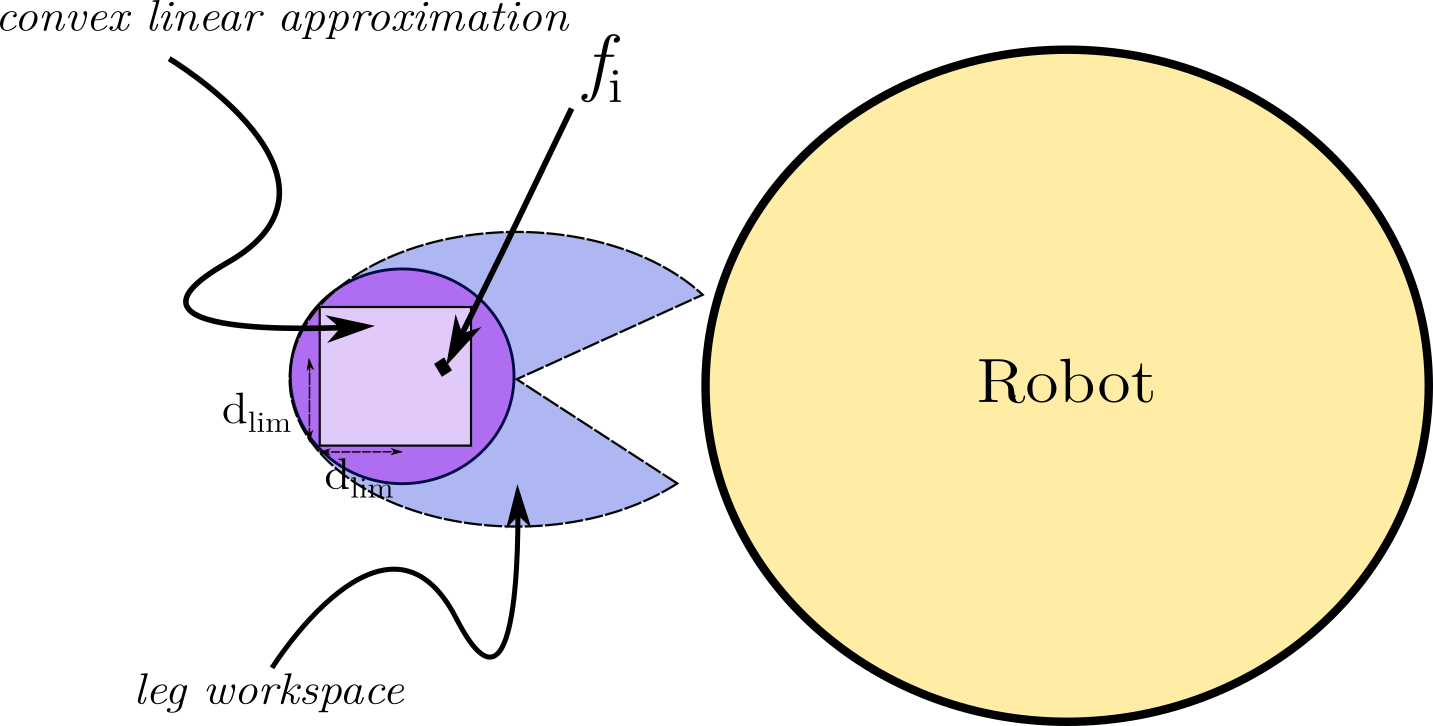}
\caption{Convex linear approximation of the kinematic reachability}
\label{fig:fig2}
\end{figure}

This allows to represent the workspace of each step in a general way that will always be convex, regardless of the configuration of the robot (as shown in figure \ref{fig:fig2}). Therefore, the following linear constraint is introduced:

\begin{equation}
f_\imath \in \begin{pmatrix} r_{nom(\imath-n_{legs}) x} \pm d_{lim} \\ r_{nom(\imath-n_{legs}) y} \pm d_{lim} \end{pmatrix}	
\end{equation}

where $d_{lim}$ represent the limits of the convex linear approximation as derived by Rojas et al. in \cite{rojas2015foothold}.

The center of the workspace is located at the foot nominal position in order to account for changes in the workspace during the base movement. Additionally, every upward and downward step is restricted to be reachable. This is represented by the following linear inequality constraint:

\begin{equation} 
| f_{\imath z} - f_{(\imath-n_{legs})z} | \ \leq \ \Delta z_{max}
\end{equation}

Where $\Delta z_{max}$ is step high variation possible.

Other formulations of reachability were also considered: Perrin et al. \cite{7750600} represents the workspace as a circle centered on nominal position of the leg, which can be introduced as quadratic constraints on the program. This approach was found to yield similar results as the presented below. However, while this representation is invariant to the rotation of the robot body it also increases the complexity of the problem by adding quadratic constraints.

\subsection{Obstacle Avoidance}

Constraining the plan to avoid obstacle collisions, or to consider uneven terrain makes the problem non-convex. A solution to this is to restrict the footstep plan to lie within a set of \textit{obstacle-free convex regions}. These regions can be easily computed by perception algorithms or by running the IRIS algorithm \cite{deits2015computing} which uses Semidefinite Programming \cite{boyd2004convex} to compute large safe convex regions in the environment. Each safe region $R$ is then represented as the following convex hull:

$$
R = \{x \in \mathbb{R}^3  \ | \ A_r x \leq b_r\}
$$

The assignment of footsteps to these regions can be done by introducing mixed-$\{0,1\}$ variables. To do this a binary matrix $\mathcal{H} \in \{ 0,1 \}^{N\times N_r}$ is defined, where $N_{r}$ is the number of safe regions. Then each footstep is assigned to a single safe region with the following linear constraint:

\begin{equation}
\sum_{r = 1}^{N_r}{\mathcal{H}_{\imath r}} = 1 
\end{equation}

While the previous doesn't mean that these regions can't intersect, it reduces the complexity of the problem by only assigning one of these regions to each step. Every footstep is then restricted to lie in a safe region by adding the following mixed-integer constraint:

\begin{equation}
\mathcal{H}_{\imath r} \Rightarrow A_r f_\imath \leq b_r 
\end{equation}

where the $\Rightarrow$ (implies) operator can be modeled in Mixed Integer Programming solvers as a linear constraint by using big-M formulation \cite{richards2005mixed}. This constraint ensures that footstep $\imath$ lies with region $r$.

\subsection{Body orientation}

\begin{figure}[ht!]
\centering
\includegraphics[width=2.5in]{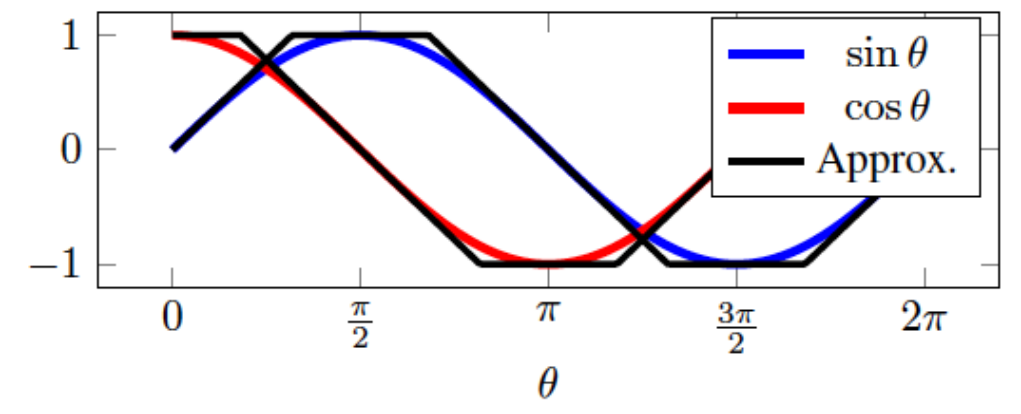}
\caption{Piecewise linear approximation of the $sin$ and $cos$ functions as shown in \cite{deits2014footstep}}
\label{fig:fig4}
\end{figure}

As previously explained, this approach uses \textit{piecewise linear approximations} of the $sin$ and $cos$ functions to maintain convexity in the problem (as shown in figure \ref{fig:fig4}). As in \cite{deits2014footstep}, two variables $s$ and $c$ are defined in order to account for $\sin(\theta)$ and $\cos(\theta)$ along with two binary matrices $\mathcal{S}$ and $\mathcal{C}$ that assign a line segment of the approximation to each variable. This is expressed with the following mixed-integer constraints:

\begin{equation}
\mathcal{S}_{\imath k} \Rightarrow \begin{cases}
\psi_{k-1} \leq \theta_\imath \leq \psi_{k+1}\\
s_\imath = m_k\theta_\imath + n_k
\end{cases}
\end{equation}

\begin{equation}
\mathcal{C}_{\imath k} \Rightarrow \begin{cases}
\gamma_{k-1} \leq \theta_\imath \leq \gamma_{k+1}\\
c_\imath = m_k\theta_\imath + n_k
\end{cases}
\end{equation}

where the angles $\psi$ and $\gamma$ represent the boundaries between each linear segment and $m$ and $n$ represent its slope and intersection. Then, it is enforced that every piecewise linear approximation lies within a single line segment, therefore for each step $f_\imath$:
	
\begin{equation}
\sum_{s = 1}^{N_s}{\mathcal{S}_{\imath s}} = 1 
\end{equation}

\begin{equation}
\sum_{s = 1}^{N_s}{\mathcal{C}_{\imath s}} = 1
\end{equation}

where $N_s$ is the total number of segments in the piecewise linear approximation.

Since the introduction of binary variables makes the problem NP-complete \cite{karp1972reducibility}, and the presence of multiple legs increases the number of segments significantly, it becomes desirable to reduce the size of the $\mathcal{C}$ and $\mathcal{S}$ matrices. The solution chosen for this is to constrain the orientation variable to be the same ($\theta$) in every $n_{legs}$ configuration of the plan, and then add a known offset when computing $\phi$ for the \textit{CoC} on each leg.\\

\subsection{Plan length}

Finally, since the necessary number of steps can't be known a priori, the chosen solution to reduce the plan size is to simply set a maximum plan length $N$, and then trim those unused steps. For this, a binary variable $t$ is introduced to determine if the step must be trimmed, and then add the following mixed-integer constraint:

\begin{equation}
t_\imath \Rightarrow f_\imath = g_{leg(\imath)}
\end{equation}

where $leg(f): \mathbb{Z} \rightarrow \{1, 2,\dots, n_{legs} \}$ is a function that returns the final step index for the leg of footstep $k$.

\subsection{Objectives}

Given the constrains stated above, the problem is formulated as a minimization program, where the objectives to minimize are:

\begin{enumerate}
	\item Final distance to the goal
	\item Number of footsteps in the plan
	\item Relative distance between configurations
\end{enumerate}

The first objective introduces a quadratic cost over the difference between the final $n_{legs}$ footsteps $f_g$ and the goal configuration $g$, formulated as:

\begin{equation}
(f_g - g)^{T}Q_{goal}(f_g - g)
\end{equation}

where $Q_{goal}$ is a weight cost matrix over the distance between each final step and the goal. Then, in order to minimize the number of footsteps in the plan it is chosen to maximize the amount of trimmed steps. Which can be achieved by assigning a negative cost over the sum of each trimming variable, and can be written as:

\begin{equation}
\sum_{k = 1}^{N}{q_t t_k}
\end{equation}

where $q_t$ is a negative weight over each trimming variable.

In order to minimize the relative distance between configurations a quadratic cost over the distance between the CoC of each configuration pair is introduced, this can be written as the following weighted sum:

\begin{equation}
\sum_{k = n_{legs}}^{N}{(p_k-p_{k-n_{legs}})^TQ_r(p_k-p_{k-n_{legs}})}
\end{equation}

where $Q_r$ is positive a weight to the cost of relative displacement between configurations.

In this implementation the biggest weight is on the distance to the goal, followed by weights the footstep trim and the relative displacement.

\subsection{Complete formulation}

The complete MIQP footstep planner with all of the constraints and objectives can be written as:

\begin{center}
\begin{tabular}{H L}
\begin{tabular}{H}
	minimize\\
	$f,t,\mathcal{H},\mathcal{S},\mathcal{C}$\\
\end{tabular}&

$
(f_g - g)^{T}Q_{g}(f_g - g) + \sum_{k = 1}^{N}{q_t t_k} + \sum_{k = n_{legs}}^{N}(p_k-p_{k-n_{legs}})^TQ_r(p_k-p_{k-n_{legs}})
$\\
\end{tabular}
\end{center}

subject to:

\begin{itemize}
\item \textit{Geometric constraints:}

$$
p_i = \frac{\sum_{k = i - n_{legs} + 1}^{i-1}{f_k}}{n_{legs} - 1}
$$

$$
r_{nom} = \begin{pmatrix} p_{x} + L_{leg} c_{\phi} \\ p_{y}  + L_{leg} s_{\phi} \end{pmatrix}
$$

$$
f_\imath \in \begin{pmatrix} r_{nomx} \pm l_{bnd} \\ r_{nomy} \pm l_{bnd} \end{pmatrix}
$$

\item \textit{XYZ reachability:}
$$
f_\imath \in \begin{pmatrix} r_{nom(\imath-n_{legs}) x} \pm d_{lim} \\ r_{nom(\imath-n_{legs}) y} \pm d_{lim} \end{pmatrix}
$$
$$
| f_{\imath z} - f_{(\imath-n_{legs})z} | \ \leq \ \Delta z_{max}
$$

\item \textit{Safe region assignment:}
$$
\sum_{r = 1}^{N_r}{\mathcal{H}_{\imath r}} = 1 
$$
$$
\mathcal{H}_{\imath r} \Rightarrow A_r f_\imath \leq b_r 
$$

\item \textit{Piecewise linear $sin$-$cos$ approximation:}
$$
\mathcal{S}_{\imath k} \Rightarrow \begin{cases}
\phi_{k-1} \leq \theta_\imath \leq \phi_{k+1}\\
s_k = m_k\theta_\imath + n_k
\end{cases}
$$
$$
\mathcal{C}_{\imath k} \Rightarrow \begin{cases}
\gamma_{k-1} \leq \theta_\imath \leq \gamma_{k+1}\\
c_k = m_k\theta_\imath + n_k
\end{cases}
$$

\item \textit{Trimming constraints:}
$$
t_\imath \Rightarrow f_\imath = g_{leg(\imath)}
$$
\end{itemize}

where $f_g = [f_N, f_{N-1}, \dots, f_{N-n_{legs}+1}]$ are the final $n_{legs} $ steps of the plan and $leg(f): \mathbb{Z} \rightarrow \{1, 2,\dots, n_{legs} \}$ is a function that receives a footstep index and returns its corresponding leg number.

\begin{figure}[h]
\centering
\includegraphics[width=2in]{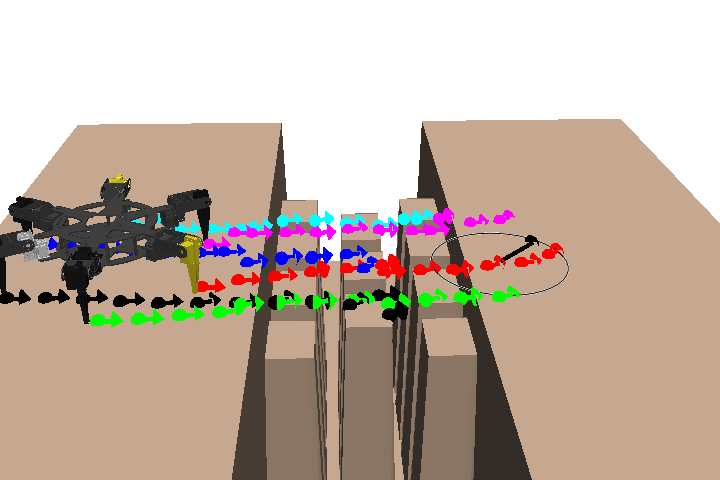}
\includegraphics[width=2in]{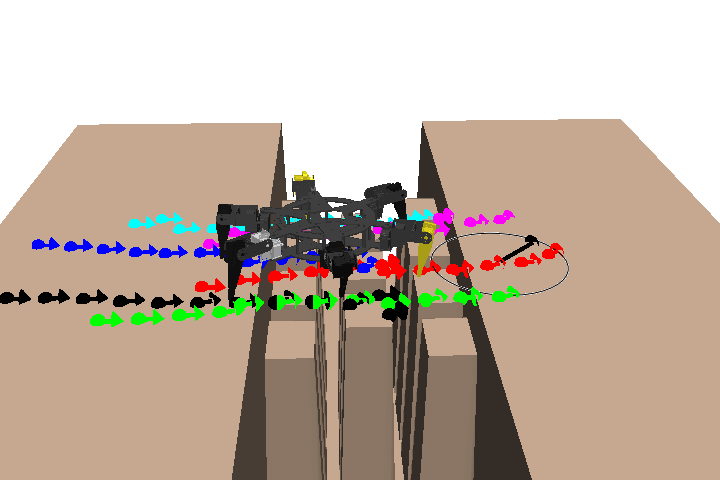}
\includegraphics[width=2in]{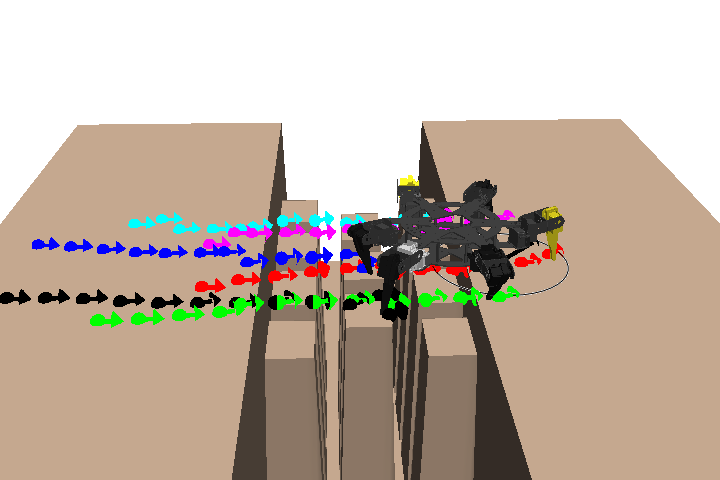}
\caption{A $BH3R$ hexapod following a generated footstep plan on a stepping stones track.}
\label{fig:fig5}
\end{figure}

\section{RESULTS}

The functionality of the planner is validated by performing simulations in MATLAB R2015a \cite{MATLAB:2015} within the Drake Toolbox for planning and control \cite{drake}, and using the optimization software Gurobi \cite{gurobi} as a solver for the MIQP.\\

Since the complexity of the problem increases significantly with the number of mixed-integer variables the tests are performed computing sets of $4 \ n_{legs}$ plans, and then concatenate them to obtain the final plan.

\subsection{Hexapods}

All tests are initially performed on a BH3R circular hexapod using a upper bound of 72 footsteps. The first test to perform is on a stepping stones track with 13 safe regions and a goal located 2 m from the initial position with a $45^{\circ}$ rotation of the base. The problem is solved returning the plan shown in figure \ref{fig:fig5}.\\

\begin{figure}[ht!]
	\centering
    \includegraphics[width=0.4\textwidth]{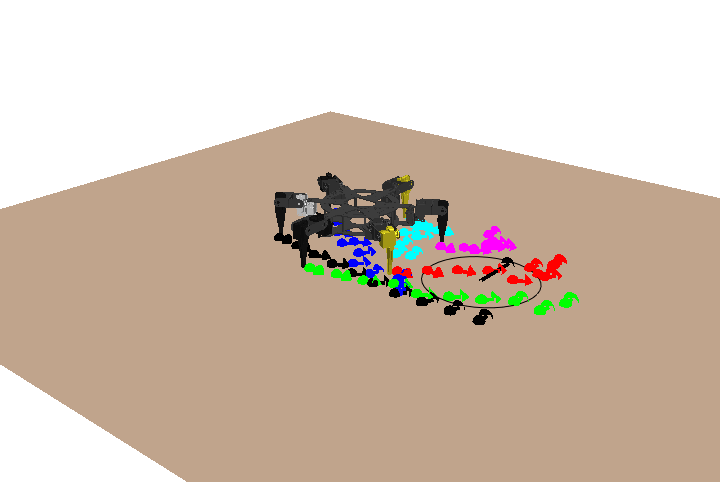}
    \caption{$BH3R$ following a footstep plan with $90^{\circ}$ rotation}
    \label{fig:fig6}
\end{figure}

\begin{figure}[ht!]
	\centering
    \includegraphics[width=0.4\textwidth]{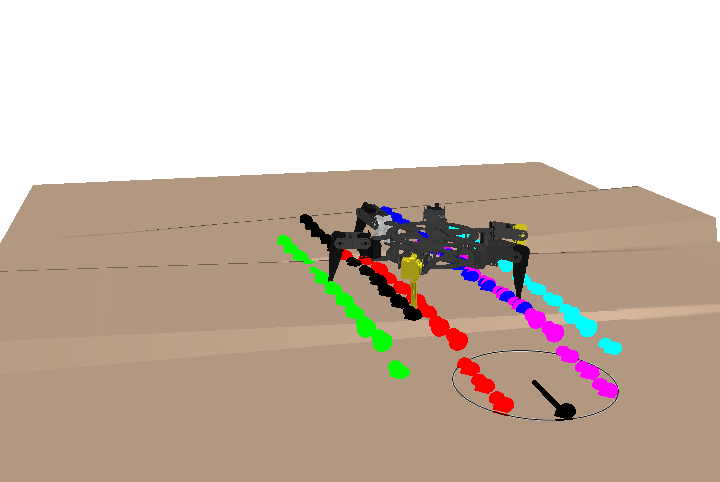}
    \caption{$BH3R$ following a footstep plan on tilted terrain}
    \label{fig:fig7}
\end{figure}

The planner is also tested with a goal rotated 90$^\circ$ from the robot, resulting in the plan shown in figure \ref{fig:fig6}. Finally the planner is tested on a tilted terrain course with a goal located 2 m from the robot, resulting in the plan shown in figure \ref{fig:fig7}.

\subsection{Quadrupeds}

To demonstrate the generality of the approach the planner is adapted to a LittleDog quadruped and perform tests on the same environments as for hexapods. Figure \ref{fig:figld1} shows the resulting plan over the tilted terrain using a 36 upper bound for footsteps.\\

\begin{figure}[t]
	\centering
    \includegraphics[width=2in]{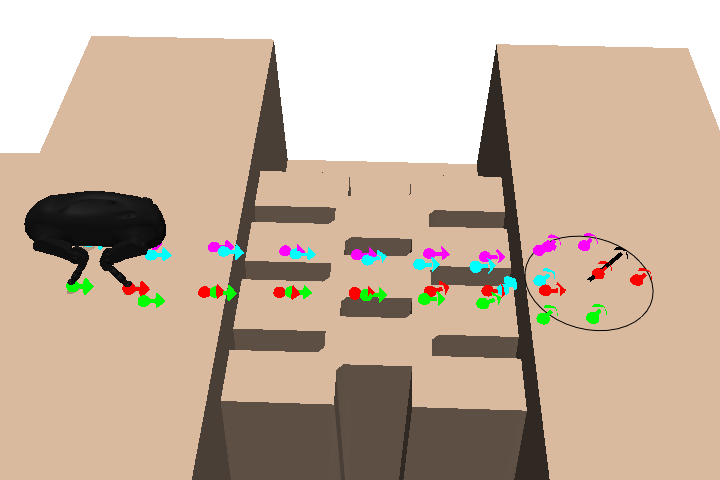}
    \includegraphics[width=2in]{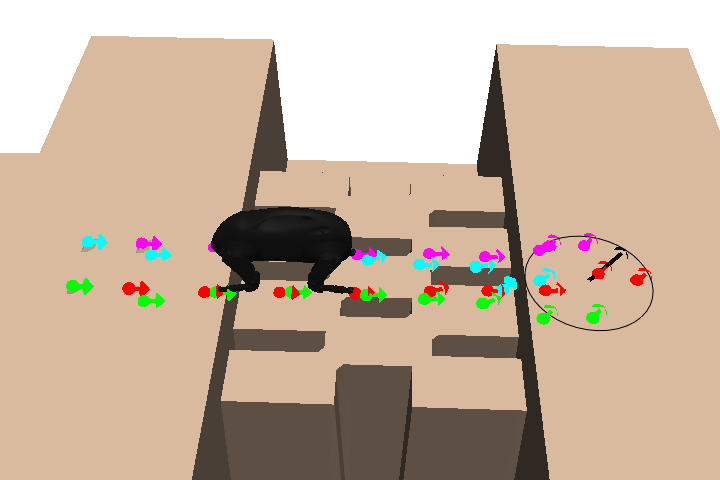}
	\includegraphics[width=2in]{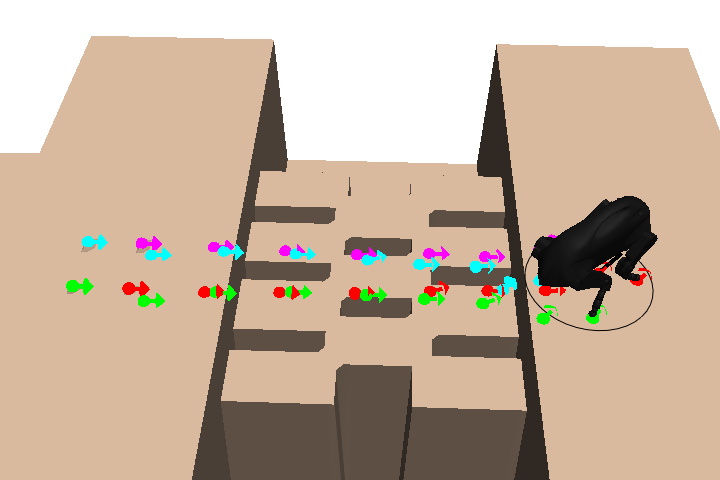}
    \caption{$LittleDog$ following a generated footstep plan for a stepping stones course}
    \label{fig:figld2}
\end{figure}

\begin{figure}[ht!]
	\centering
    \includegraphics[width=0.4\textwidth]{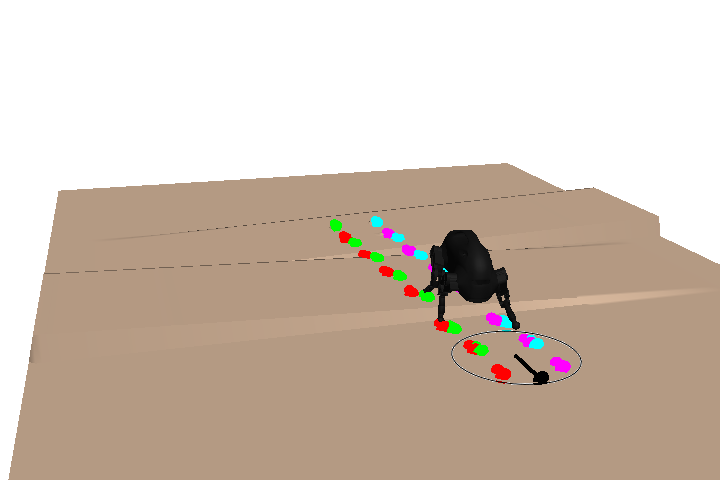}
    \caption{$LittleDog$ following a generated footstep plan for a tilted terrain course}
    \label{fig:figld1}
\end{figure}

Then a planner is tested with a tilted terrain course, using the same specifications as before with a goal located 2 m ahead of the robot. A plan is generated and is shown in figure \ref{fig:figld2}.

\subsection{Performance}

All of the tests were performed on a low-end commercial 2.4 GHz \textit{Intel Core 2 Quad} computer with Ubuntu 15.04, similar to \textit{onboard processors}. As noted above the complexity of the problem increases significantly with the number of integer variables on the problem and therefore it can result more efficient to plan considering fewer safe regions or smaller rotation ranges.

\begin{table}[ht!]
	\centering
    \caption{Performance of the MIQP for different scenarios}
    \label{tab:tab1}
    \begin{tabular}{|c|c|c|}
	\hline
    Footsteps& Mixed-Integer Variables & Solving Time (s)\\
   	\hline
    36& 828 & 3.68 \\
    \hline
    24& 552 & 0.44 \\
    \hline
    12& 312 & 0.08 \\
    \hline
	\end{tabular}
\end{table}

To contrast the impact this results have in the performance of the planner Table \ref{tab:tab1} presents a comparison of the solving time in different scenarios were the planner is challenged. It is important to note the variation in the complexity of the problem when the number of mixed-integer variables increases.

\section{CONCLUSION}

This work introduces a generalized continuous optimization approach to footstep planning; the proposed approach leverages Mixed-Integer Programming to replace non-convex constraints and can be easily adapted to multilegged robots with different geometries. This approach was successfully implemented on Drake \cite{drake} and represents a successful implementation of a generalized continuous optimization footstep planner for different multilegged robots.

The results obtained show how this approach handles complex rotations successfully; even on cluttered environments and rough terrain without a significant increase of the complexity of the problem. The raw planner can return average sequences of footsteps in matter of seconds, on easy mockups; it can return solutions in few minutes when the planning is done on complex environments. When concatenating shorter plans it can be solved for the same sequences, in fragments of a second, using low-end processing.

\subsection{Future work}

The performance of the planner represents an opportunity to develop \textit{online footstep planners} based on continuous optimization by working with short sequences of footsteps and recomputing offboard for changes in the environment, similar to the work shown in \cite{karkowski2016real}. This would require a set of local goals in order to avoid local minima (for instance, a wall between the robot and the goal), possibly by running a discrete search over a sampled space of the convex safe regions using RRT* or ARA*.

Another opportunity is to integrate novel robust walking planning approaches on multilegged platforms. New optimization-based approaches have been successfully implemented for humanoid platforms using robust optimization to maximize a universal stability margin based on \textit{Contact Wrench Cones} as shown in \cite{daiplanning}.

Furthermore, the planner itself still has room for improvement, since the workspace of a multilegged robot changes significantly when performing dynamic motions. So, it becomes necessary to account for these changes when representing the geometric constraints. Adding this consideration and including footstep sequence within the planner, assigning a binary variable to the moving legs, would allow to obtain different locomotion modes and gaits.

\subsection{Source code}

Following the same spirit as \cite{deits2014footstep} the authors of this work have made the entire source code publicly available on github\footnote{\href{https://github.com/baceituno}{https://github.com/baceituno}}.

\section*{ACKNOWLEDGMENT}

The authors would like to thank Hongkai Dai for his advise and expertise in the subject, and Maureen Rojas for her support on the simulations stage of this project. This work was partially funded by the Universidad Sim\'on Bol\'ivar Research and Development deanship.

\bibliographystyle{IEEEtran}
\bibliography{IEEEabrv,references}

\end{document}